\documentclass[conference]{IEEEtran}
\usepackage{ifpdf}

\usepackage{cite}

\ifCLASSINFOpdf
  \usepackage[pdftex]{graphicx}
   \graphicspath{{../pdf/}{../jpeg/}}
   \DeclareGraphicsExtensions{.pdf,.jpeg,.png}
\else
   \usepackage[dvips]{graphicx}
   \graphicspath{{../eps/}}
   \DeclareGraphicsExtensions{.eps}
\fi
\usepackage{algorithmic}

\usepackage{array}

\ifCLASSOPTIONcompsoc
 \usepackage[caption=false,font=normalsize,labelfont=sf,textfont=sf]{subfig}
\else
 \usepackage[caption=false,font=footnotesize]{subfig}
\fi
\usepackage{fixltx2e}
\usepackage{url}

\usepackage{amsmath}
\usepackage{amssymb}
\usepackage{url}

\usepackage{graphicx}
\usepackage{xcolor}
\usepackage{epstopdf}
\graphicspath{ {fig/} } 
\usepackage{multirow}
\usepackage{rotating}
\usepackage{diagbox}
\usepackage{setspace}
\usepackage[colorlinks,linkcolor=blue,bookmarks=false]{hyperref}

\usepackage{floatrow}
\DeclareFloatFont{tiny}{\tiny}
\usepackage{lipsum}
\usepackage{dblfloatfix}

\hypersetup{draft}

\begin{document}

%
\title{A Wireless-Vision Dataset for Privacy\\Preserving Human Activity Recognition}

\author{\IEEEauthorblockN{Yanling Hao}
\IEEEauthorblockA{Queen Mary University of London\\
London, UK\\
Email: yanling.hao@qmul.ac.uk}
\and
\IEEEauthorblockN{Zhiyuan Shi}
\IEEEauthorblockA{Onfido Research London\\
London, UK\\
Email: zhiyuan.shi@onfido.com}
\and
\IEEEauthorblockN{Yuanwei Liu}
\IEEEauthorblockA{Queen Mary University of London\\
London, UK\\
Email: yuanwei.liu@qmul.ac.uk}}


%


\maketitle

\begin{abstract}
Human Activity Recognition (HAR) has recently received remarkable attentions in numerous applications such as assisted living and remote monitoring. Existing solutions based on sensors and vision technologies have obtained achievements but still suffering from considerable limitations in the environmental requirement. Wireless signals like WiFi-based sensing has emerged as a new paradigm since it is convenient and not restricted in the environment. In this paper, a new WiFi-based and video-based neural network (WiNN) is proposed to improve the robustness of activity recognition where the synchronized video serves as the supplement for the wireless data. Moreover, a wireless-vision benchmark (WiVi) is collected for 9 class actions recognition in three different visual conditions,  including the scenes without occlusion, with partial occlusion, and with full occlusion.  Both machine learning method - support vector machine (SVM) as well as deep learning methods are used for the accuracy verification of the data set. Our results show that WiVi data set satisfies the primary demand and all three branches in proposed pipeline keep more than $80\%$ of activity recognition accuracy over multiple action segmentation from 1s to 3s. In particular, WiNN is the most robust method in terms of all the actions on three action segmentation compared to the others. 

\end{abstract}

%
\IEEEpeerreviewmaketitle

\section{Introduction}
Human activity recognition has emerged as an important task recently in numerous applications,  such as assisted living~\cite{wu2018wifi}, human-computer interaction~\cite{zou2018towards}, health monitoring~\cite{tan2018exploiting}, surveillance~\cite{wang2018csi}, etc.  In existing systems, the individual has to wear a device equipped with motion sensors such as a gyroscope and an accelerator. The sensor data is processed locally on the wearable device or transmitted to a server for feature extraction, and then supervised learning algorithms are used for classification. This type of monitoring is known as active monitoring. The performance of such a system is shown to be around 90 percent for recognition of activities such as sleeping, sitting, standing, walking, and running. However, it is not always convenient to wear devices to monitor the activities in passive applications because of the additional burden and discomfort associated with wearing the device.

Camera-based systems play an important role in passive activity recognition. However, occlusion remains a fundamental challenge. The requirement for light is a major limitation for such systems. Furthermore, it often refers to privacy issues and is limited in many conditions. Unlike sensor-based and video-based solutions, WiFi sensing is not intrusive without the privacy issue and insensitive to lighting conditions. Therefore, WiFi sensing has obtained prevailing popularity in human action recognition~\cite{zou2018wifi}. 

Recent years have witnessed rapid growth of techniques using channel state information (CSI) of WiFi signals in sensing human bodies~\cite{wang2018wi,wu2018tw}. Although WiFi CSI  has obtained much progress in localizing people and tracking their motion, it is cumbersome for fine-grained human activities recognition because CSI is sensitive to environmental influences.  Moreover, the vision information from the camera is a potential complement to assist CSI in some environments without any occlusions. These questions, give us an incentive to develop a new WiFi and video combined pipeline (WiNN) for human action recognition. The synchronous video and WiFi CSI with common supplement and mutual complementary makes WiNN adjustable and feasible for real conditions, especially complex visual conditions, including the scenes without occlusion, with partial occlusion, and with full occlusion, respectively.

To summarize, the contributions of this paper are three-fold.

• We first construct WiVi, wireless-vision activity data set, as a benchmark to evaluate the performance of existing activity recognition systems. We tested SVM,  convolution neural network (CNN) as well as the proposed WiNN to verify the effectiveness of the WiVi dataset.

• We propose WiFi-based and video-based neural network as WiNN for activity recognition in scenes with partial occlusion and with full occlusion, which improves the robustness of activity recognition according to the synchronous video as a supplement and complement for WiFi CSI signals.

• We compare machine learning method SVM and deep learning methods CNN and WiNN to verify the quality of the WiVi  data set. Particularly,  WiNN achieves the most robust results over multiple action segmentation from 1s to 3s.

\section{Related work}
\begin{figure*}[t]
    \centering
    \includegraphics[width=\textwidth]{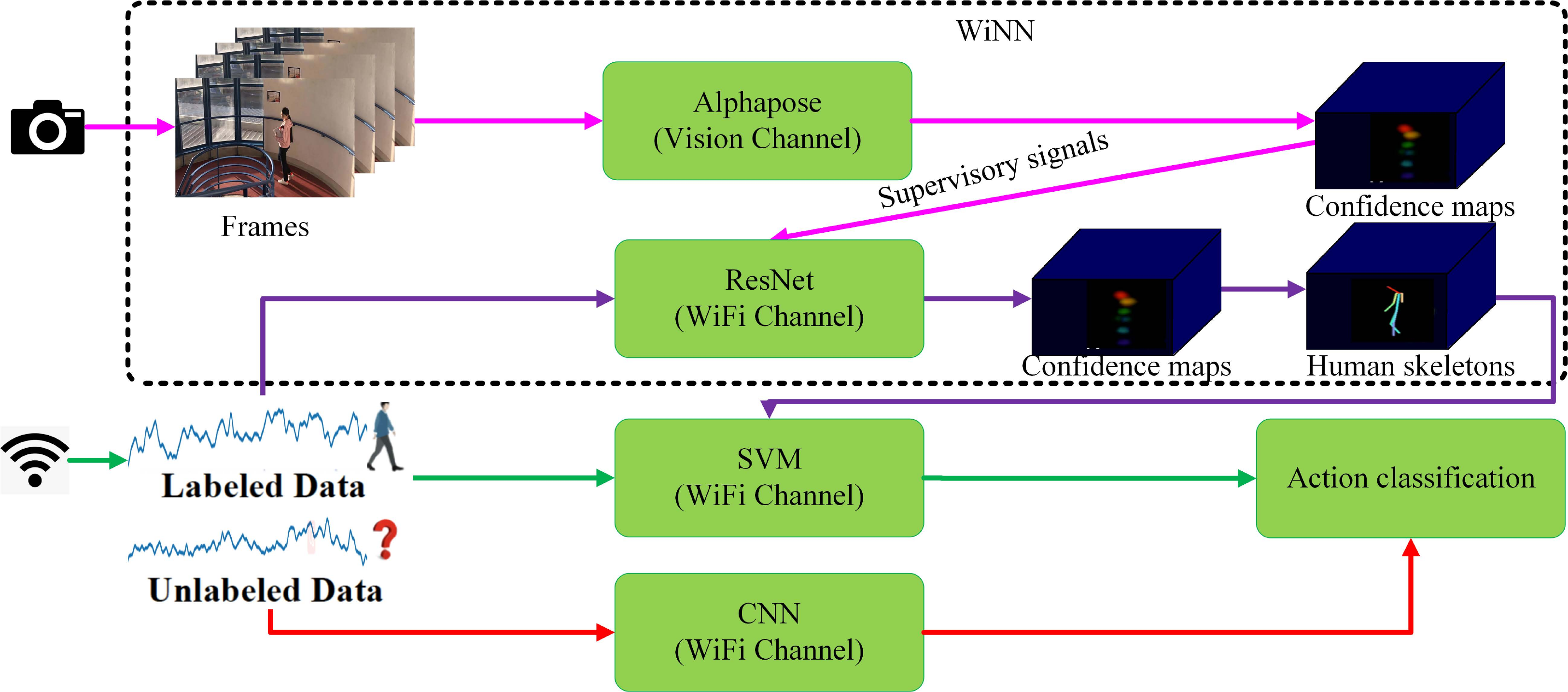}
    \caption{The flowchart of our network for WiVi dataset.}
    \label{fig:pipeline}
\end{figure*}

To detect human action, two major solutions have been paid attention, namely, camera-based methods~\cite{fang2017rmpe,cao2017realtime} and wireless signal based methods including radars (depth maps)~\cite{zhao2018through}, LiDARs (3D point clouds)~\cite{maturana2015voxnet,garcia2016pointnet} and WiFi devices~\cite{pu2013whole,qian2017inferring}. 

\subsection{Camera-based Methods}

Camera-based solutions often generate 2D RGB or depth images or videos.  It has got more research attention and been prevailing in human recognition. Traditionally, camera-based methods approach detect human activities by using either a two-step framework~\cite{ren2015faster,liu2016ssd,redmon2016you,lin2017feature} or a part-based framework~\cite{chen2015parsing,insafutdinov2017arttrack}. The two-step framework first detects all humans forming the bounding box, and then estimates the action from the bounding box independently. The part-based framework first detects body parts independently and then assembles the detected body parts to form multiple human poses to recognize the actions. Both frameworks have their advantages and disadvantages. Yet, camera-based recognition methods are prone to be influenced by the environment such as background, lighting and occlusion, and social limitations such as privacy concerns.

\subsection{Wireless Signal based Methods}

The use of passive radar and LiDARs sensors based non-cooperative detection is of increasing interest in many application areas, including defense, security, transport, and healthcare. However, these methods often require dedicated and specially designed hardware. 

WiFi devices, which are cheaper and power-efficient than radars and LiDARs, invariant to illumination, and have little privacy concern comparing to cameras. However, thus far the potential of WiFi-based methods has not been fully exploited. Mostly WiFi-based methods can only generate a coarse resolution of the recognition and thereby often are used for the perception such as indoor localization~\cite{adib2013see} and the rough classification~\cite{ali2015keystroke,li2016csi,qian2017inferring}. To increase the robustness, some research adopts the WiFi arrays to simulates a 2D antenna array to improve the accuracy of recognition and localization ~\cite{huang2014feasibility}. The authors in ~\cite{holl2017holography}  have made an attempt to use the WiFi CSI information combined with the videos for human pose interpretation. He adopted Alphapose to generate the 2D skeleton of human pose from the video as the supervised learning of a deep neural network first and then get the human pose estimation by using the WiFi CSI only. Inspired by this two branch idea, we propose a four-branch pipeline for human action recognition.

\section{Proposed WiFi- and Video- based Pipeline}

\subsection{WiFi-based and Video-based Neural Networks}
Convolution neural network (CNN) is a kind of multi-layer neural network, which can extract depth features hierarchically from higher resolution and then convert them into more complex low-resolution features. The classical convolutional neural network consists of two trainable network structures: the convolutional layer, and the pooling layer. The convolutional layer and the pooling layer are used as feature extraction layers, where the convolutional layer provides filtering generation of convolution feature mapping, and the pooling layer produces more abstract and robust high-level features from the convolution feature. The output layer is the full connection layer, dedicated to sorting tasks. 

In this work, we design CNN and WiNN based on convolutional neural networks as deep learning methods for the CSI classification. As shown in Fig. \ref{fig:pipeline}. The whole scheme is composed of three branches including WiNN (in purple lines), SVM (WiFi channel drawn in green lines), and CNN (WiFi channel with red lines). In addition, all of them can realize HAR based CSI signals. As seen in  Fig. \ref{fig:pipeline}, the SVM and the CNN branches apply classification on WiFi CSI data only. By contrast, the WiNN branch uses the video as complementary for WiFi signals. WiNN contains two channels, namely the vision channel and the WiFi channel.  The vision channel adopts the Alphapose \cite{fang2017rmpe}, which takes the videos in the scene without occlusion as input in order to generate the human pose skeletons. The generated skeleton results further serve as the supervision for the WiFi channel learning. Then, the WiFi channel trains the network by the CSI samples according to the supervising. The trained WiFi channel is able to generate the human skeleton based on the CSI data only without the video. Based on the trained network, WiNN has the ability to generate the human pose skeleton results from CSI samples in occlusion conditions, and even dark scenes. Finally, the generated skeletons act as the input of the SVM classifier for action classification. 

\begin{figure*}[t]
    \centering
    \includegraphics[width=\textwidth]{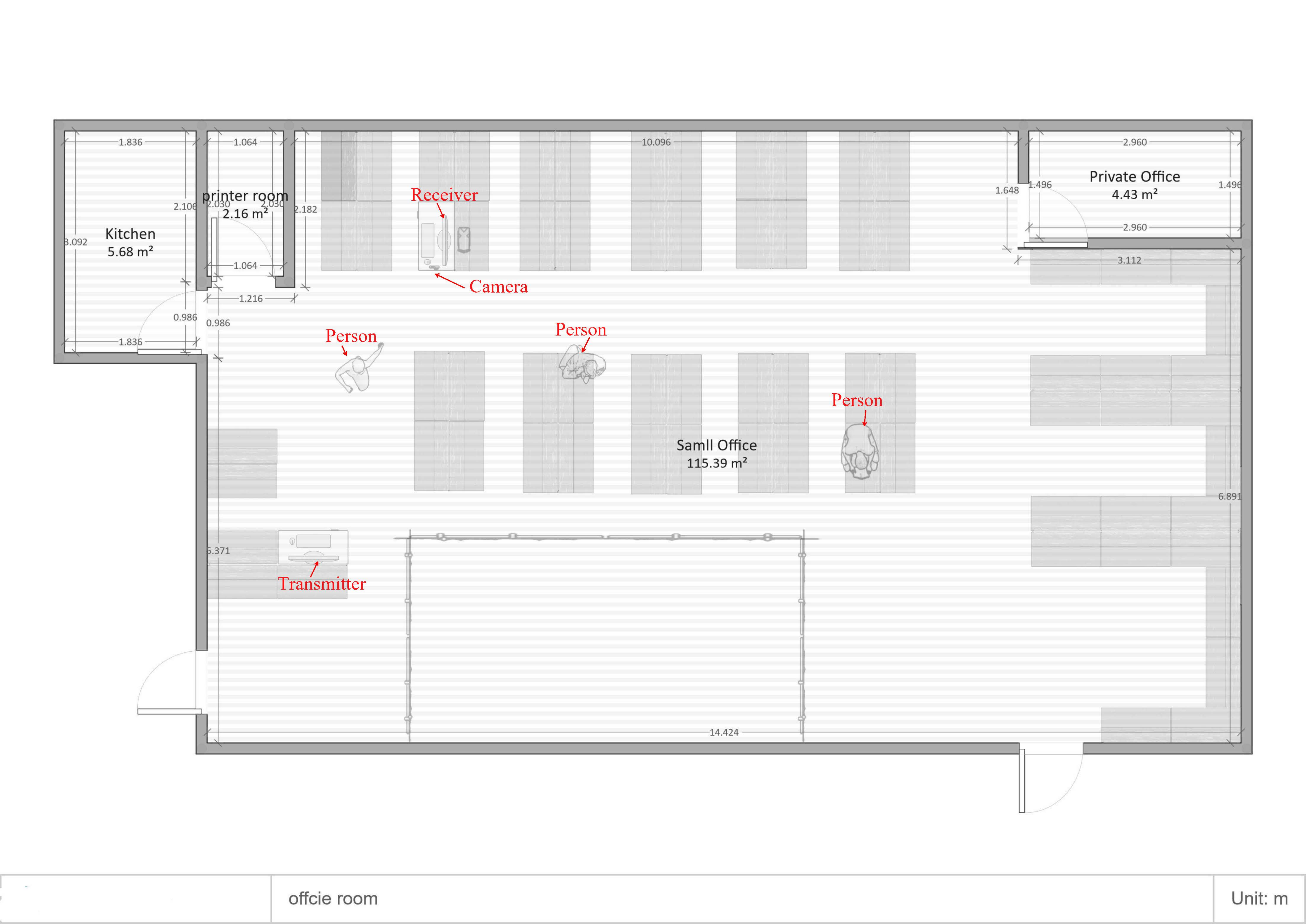}
    \caption{The floor plan of the experiment scene. The transmitter desktop and the receiver desktop are fixed located in the left side of the office, while the volunteer persons perform activities in different areas to simulate the occlusion cases }
    \label{fig:scenariopic}
\end{figure*}

\subsection{Channel State Information}
CSI depicts how a signal propagates from the transmitter to the receiver at a certain scale of subcarriers frequencies along multiple paths. Generally, it can be modeled as follows:
\begin{equation}
\boldsymbol Y=\boldsymbol H\times\boldsymbol X+\boldsymbol N
\end{equation}
where $\boldsymbol X $ and $\boldsymbol Y$ correspond to the transmitted and received signal vectors, respectively. $\boldsymbol H $ is the channel matrix representing the CSI values, and $\boldsymbol N$ is the additive white Gaussian noise vector. 
In the frequency domain, CSI defines the responses derived from orthogonal frequency-division multiplexing (OFDM) subcarriers, which can be defined as 
\begin{equation}
h=\left|h\right|e^{j\sin\theta}
\end{equation}
where $\left|h\right|$ and $\theta $ are the amplitude and phase. $ h $ defines the CSI values of each subcarrier. In this article, CSI contains 30 subcarriers to describe the multipath propagation affected by the physical environment (e.g., reflection, diffraction, and scattering). In particular, CSI amplitudes are further used for analysis  with respect to the fact that phase information are not generally used for action recognition due to carrier frequency offset (CFO) and sampling frequency offset (SFO) errors. We activate 3 antennas both in transmitter and receiver, and then each obtained CSI data packet contains up to $ {\rm{3}} \times {\rm{3}} \times {\rm{30 = 270 }} $  raw features of amplitudes. 
\begin{figure}[!t]
    \centering
    \includegraphics[width=\textwidth]{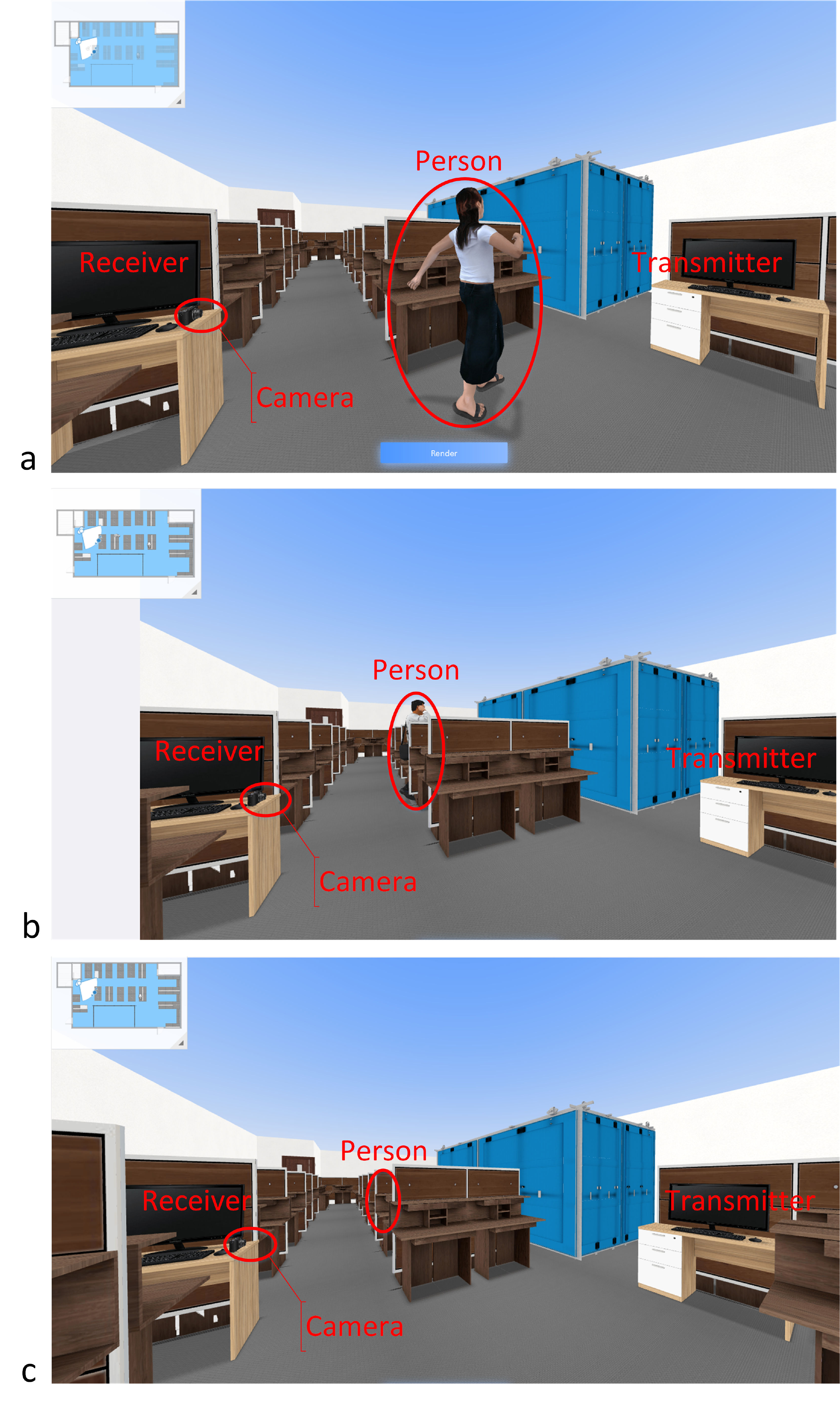}
    \caption{Three scenarios design, including the scenes a. without occlusion, b. partially occlusion and c. fully occlusion}
    \label{fig:scenedesign}
\end{figure}

\subsection{Constructing Wireless-vision Dataset (WiVi)}

The data collection is conducted in one spacious office apartment with five different rooms, i.e. kitchen, printer room, private office, and the main office room as shown in Fig. \ref{fig:scenariopic} . The office rooms have standard furniture: tables, chairs, boards, etc.  The interior walls of the building are 6-inch hollow walls supported by steel frames with sheetrock on top.  The door and windows were closed to ensure stability in the test environment.

The experimental hardware consists of two desktop computers as transmitter and receiver operating in IEEE 802.11n monitor mode at 5.4 GHz. Both desktops are equipped with off-the-shelf Intel 5300 card. They run Ubuntu 14.04 with a modified Intel driver. During the experiment, the transmitter desktop and the receiver desktop are fixed located on the left side of the office, while the volunteer persons perform activities in different areas to simulate the occlusion occasions. 

During the data capturing,  wireless and vision data is synchronously collected. WiFi signals are collected by the receiver at a sampling rate of 100 packets per second. The receiving desktop computer calculates CSI data for every packet received from the transmitter. We employ the CSI extraction tool (introduced in \url{https://github.com/dhalperi/linux-80211n-csitool-supplementary}) for CSI signals recording and CSI packets extraction. The CSI measurement from each data packet contains up to $ {\rm{3}} \times {\rm{3}} \times {\rm{30 = 270 }} $  raw features. 

To keep pace with the CSI recording,  a deep camera D435i (available at \url{https://www.intelrealsense.com/depth-camera-d435i/} ) is attached to our receiver desktop at the same location of the wireless card. To synchronize the images and wireless data, we record the video at 20 FPS. That is, every five CSI  samples are corresponding to one frame in the video, as shown in Fig. \ref{fig:falling}. Fig. \ref{fig:falling}a is the CSI sequence and Fig. \ref{fig:falling}b is the corresponding frame recorded in videos.

We recruited 2 volunteers to join our evaluation, including 1 male and 1 female. We implemented 9 activities, i.e. falling down, throwing, pushing, kicking, punching, jumping, drinking, phone talking, and seating. 

In reality, such kind of activities is complex. To simulate as many situations as possible, each action is repeatedly performed by the volunteer and lasts about 20-30 seconds. During the time, the activities are repeated in different motion details, such as varied directions, speed, intensity, strength, and etc., to ensure the diversity of the actions. In addition, we also simulate the occlusion occasions, as shown in Fig. \ref{fig:scenedesign}, including three scenarios,  namely, the scenes without occlusion (as shown in Fig \ref{fig:scenedesign}a), with partially occlusion ( as seen in Fig. \ref{fig:scenedesign}b) and with fully occlusion (Fig. \ref{fig:scenedesign}c).

\subsection{Pre-processing of CSI}
\begin{figure*}[t]
    \centering
    \includegraphics[width=\textwidth]{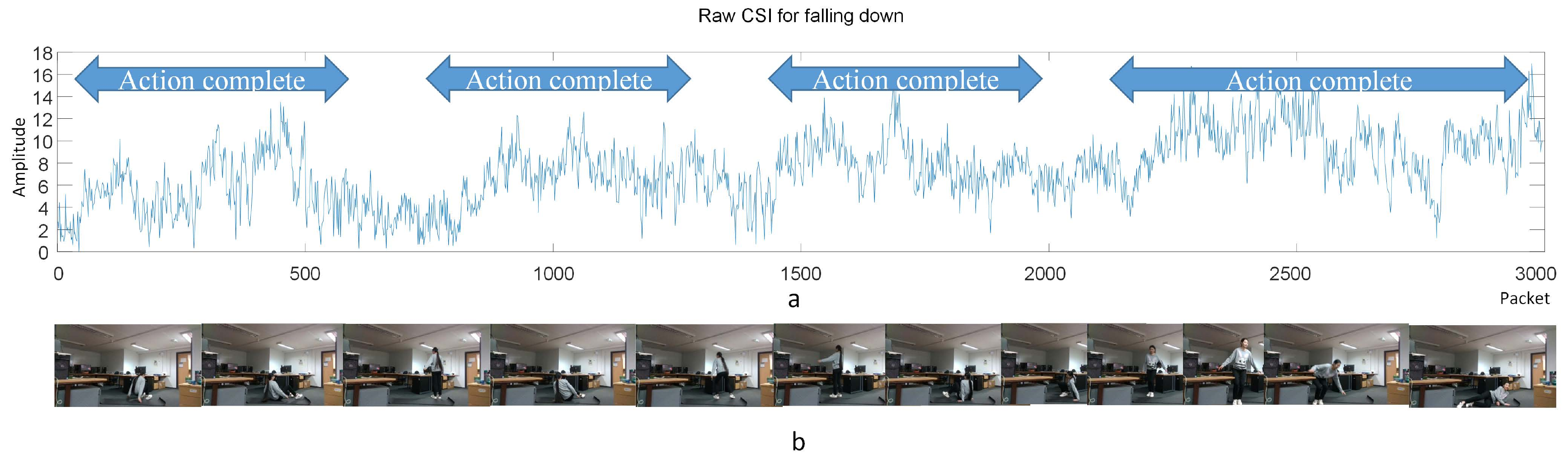}
    \caption{Raw CSI packets and synchronous frames of female person for falling down in the scene without occlusion. The data is CSI in the first subcarrier between the first transmitter antenna and the first receiver antenna. According to frame images, the complete action is denoted for the CSI sequence.}
    \label{fig:falling}
\end{figure*}

\begin{figure*}[t]
    \centering
    \includegraphics[width=\textwidth]{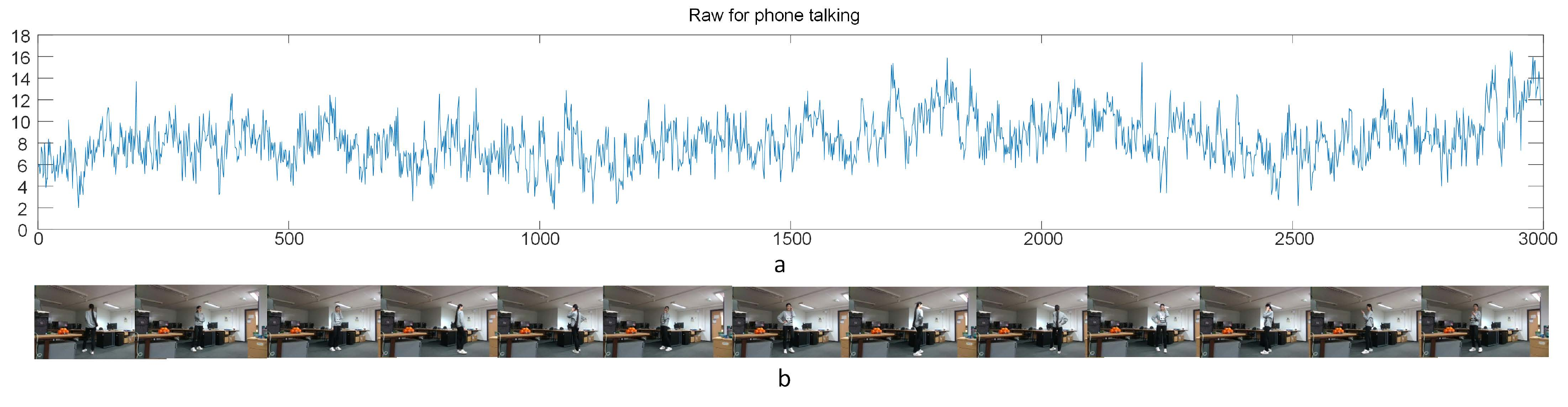}
    \caption{Raw CSI packets and synchronous frames of female person for phone talking in the scene without occlusion. The data is CSI in the first subcarrier between the first transmitter antenna and the first receiver antenna. }
    \label{fig:talking}
\end{figure*}

The gathered CSI is easily affected by the surrounding electromagnetic noise. An example for dynamic falling-down action is demonstrated in Fig.\ref{fig:falling}a. A relatively static phone-talking action is shown in Fig.\ref{fig:talking}a. Both of them are from the first subcarrier between the first transmitter antenna and the first receiver antenna of the female person in the scene without occlusion. Fig.\ref{fig:falling}a and Fig.\ref{fig:talking}a show that the CSI sequence both behaves periodicity changes of signals when the person performs an action. Meanwhile, the fluctuations of the dynamic and the static action curves are obviously different. Therefore, the CSI has the potential to distinguish different actions. However, both of the two action CSI  sequences are very noisy signals. To address the issue, we apply three pre-processing steps to remove the noises.

\begin{figure}[t]
    \centering
    \includegraphics[width=\textwidth]{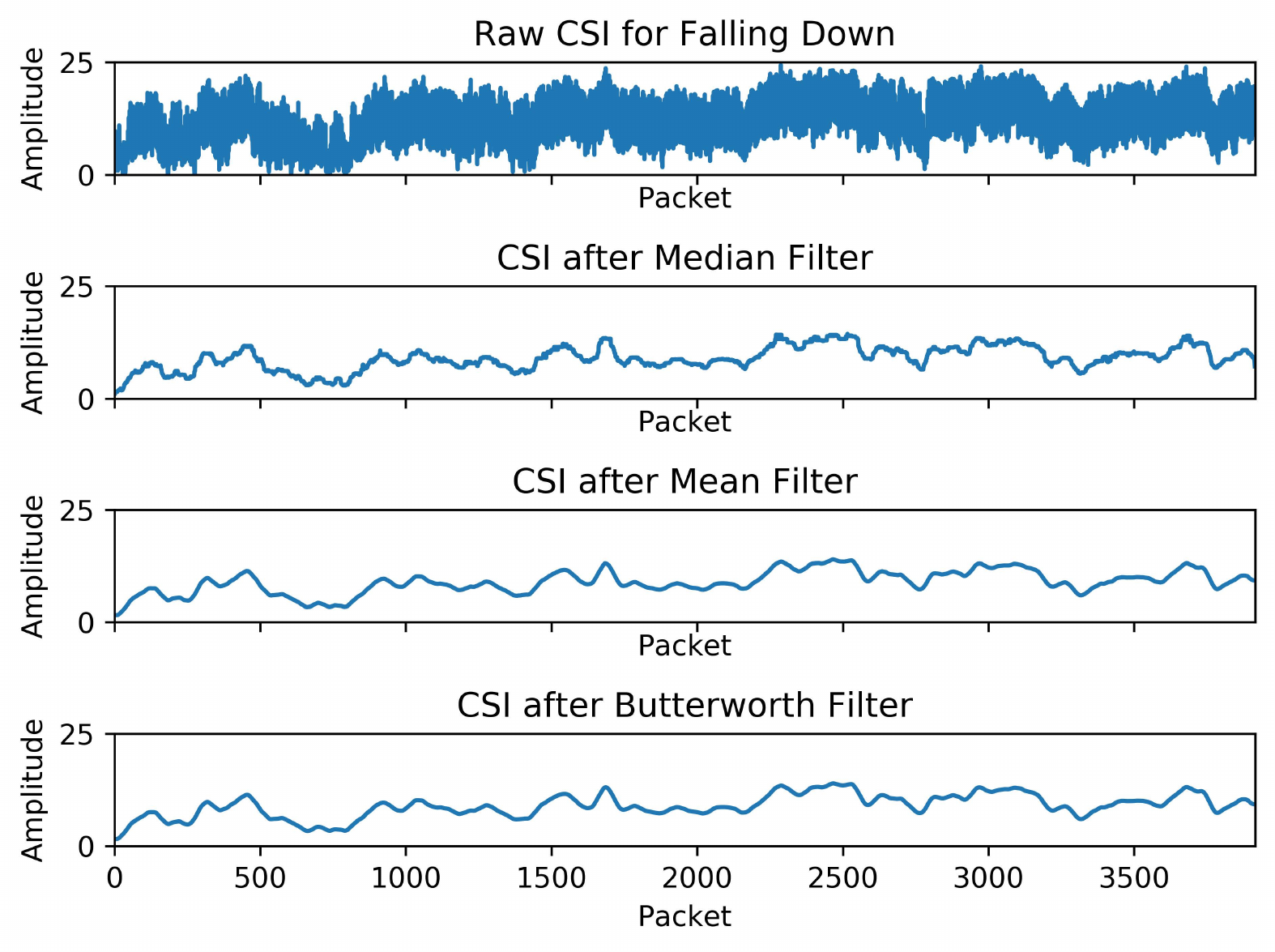}
    \caption{Raw CSI packets, CSI after median fiter, CSI after mean filter and CSI after Butterworth filter for falling down from the first subcarrier between first transmitter antenna and first receiver antenna in the scene with occlusion.}
    \label{fig:falling_f}
\end{figure}

\begin{figure}[t]
    \centering
    \includegraphics[width=\textwidth]{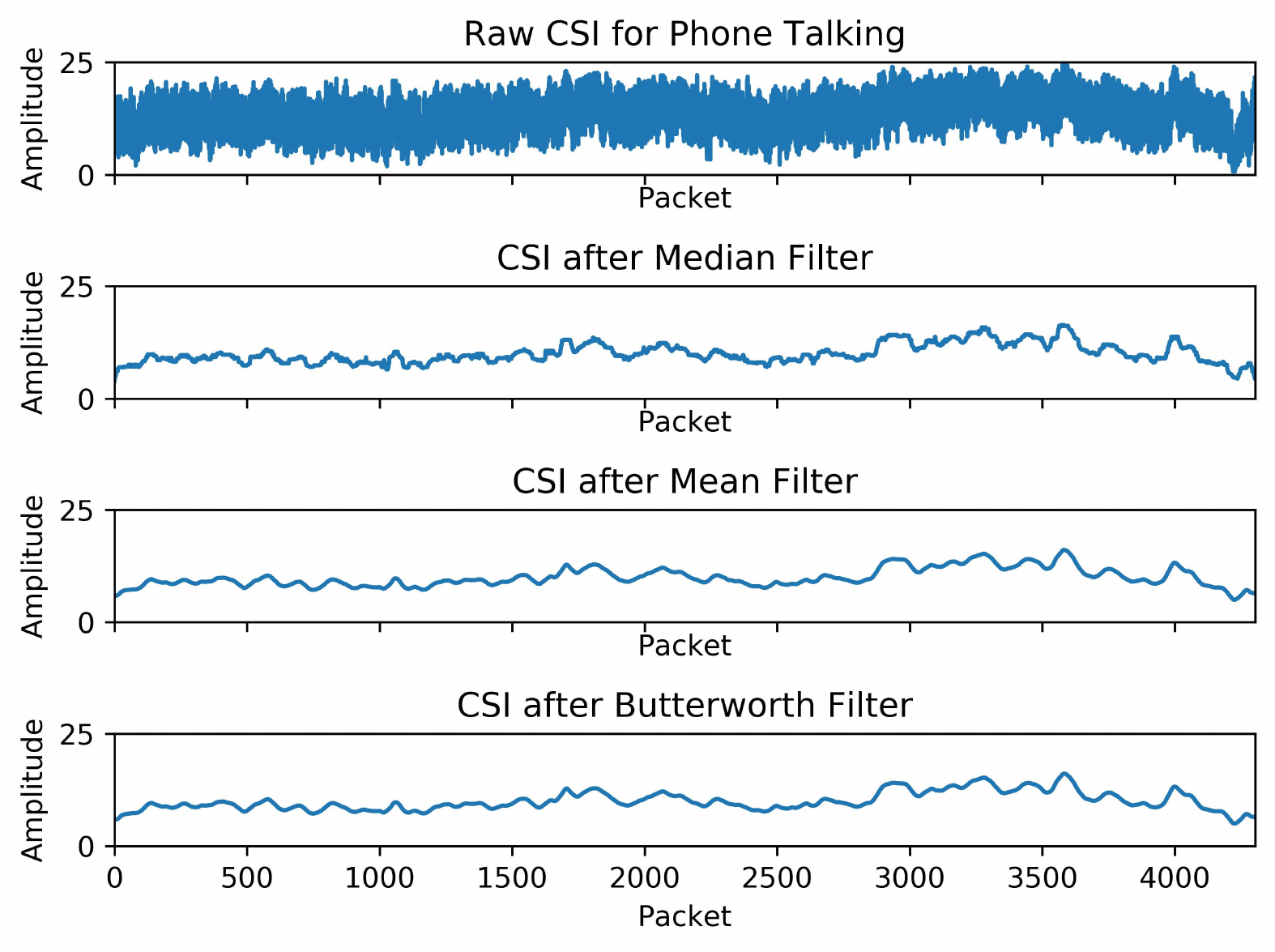}
    \caption{Raw CSI packets, Raw CSI packets, CSI after median fiter, CSI after mean filter and CSI after Butterworth filter for phone talking from the first subcarrier between first transmitter antenna and first receiver antenna in the scene with occlusion.}
    \label{fig:talking_f}
\end{figure}

\begin{itemize}
\item Outliers removing.  As seen in the first rows of Fig. \ref{fig:falling_f} as well as Fig. \ref{fig:talking_f}, the raw CSI inevitably contains some outliers, that values are extremely high or low. To mitigate the performance of CSI,  a median filter is applied to get a smooth CSI sequence, where the sliding window is 40, as illustrated in the second row of Fig. \ref{fig:falling_f} as well as Fig. \ref{fig:talking_f}.
\item Smoothing. After the above operation, there still exists slight fluctuations, the median filter is adopted to smooth the CSI sequences with a sliding window of 40. The third rows of Fig. \ref{fig:falling_f}  and  Fig. \ref{fig:talking_f} draw the smooth results. 
\item Low pass filtering. Butterworth filter with fifth order and  10Hz pass-band is implemented to filtered out high-frequency noise, as shown in the fourth rows Fig. \ref{fig:falling_f} and Fig. \ref{fig:talking_f}.
\end{itemize}

\subsection{Action Segmentation and Data Augmentation}
A prerequisite for further CSI processing such as classification is action segmentation, as the CSI sequences are recorded along time.  Ever since the advent of CSI, tremendous efforts have been made to develop robust and accurate action segmentation, such as the classical envelope extraction method~\cite{huang1998empirical} by decomposing data into intrinsic mode functions for action separation. However, this method is only viable supposing the scenarios is stationary and not suitable in most actual cases. In addition, there are differences varying from person to person causing the variance in the movement. Therefore the empirical design of a special action segmentation method is cumbersome for other cases.  How to find an effective way of action segmentation remains a fundamental challenge. To this end, we propose a way for action segmentation by time in this work. That is because each CSI sequence records the same activities, it is assumed to be separated into different movements by an equal time interval.  

After the action segmentation, to make CSI actions set more robust to the noise, data augmentation is performed on the action sets. Inspired by the data augmentation method in ~\cite{wang2018csi}, we propose a method including two steps also.  
\begin{itemize}
\item Averaging an action into CSI samples. We first uniformly split the sequence into segments as the individual actions. Then we extract 10 CSI packages with equal intervals in a segment derived from the action segmentation step, and then iteratively select every 10 CSI packages with a stride of size 1. At last, these 10 CSI packages are averaged into one CSI sample separately for further processing.
\item Data augmentation for the training data. With respect to the limited number of CSI sequences, it tends to cause the over-fitting problem. To address this obstacle, the data augmentation described in~\cite{wang2018csi} is applied with $k={3, 5, 7, 9}$ to augment the dataset.
\end{itemize}

\section{Implementation and Evaluation}
\subsection{Experiment Setup}
\subsubsection{Datasets}
As described in subsection B of section IV, we use 1, 2, 3 seconds to segment the CSI sequences to get the actions dataset WiVi. Then, we average the action segmentation sets into CSI samples and augment the CSI samples to form the training datasets WiVi. At last, the dataset for three action segmentation are shown in Table.\ref{tab:datasets}. The dataset was split into a training and a testing sets with an {\rm{70 - 30\% }} split respectively. All the test results on our dataset use only WiFi signals without vision-based input.

\begin{table}[h]
\caption{The dataset for three action segmentation}
\label{tab:datasets}
\resizebox{\textwidth}{!}{
\begin{tabular}{ccc}
\hline
\diagbox{Segment time}{Sets}
 & training instances & testing instances \\
\hline
1s   & 24938              & 10707             \\
2s   & 24398              & 10483             \\
3s   & 23840              & 10280             \\
\hline
\end{tabular}}
\end{table}

In this section, we employed both the traditional machine learning and deep learning methods to verify the efficiency of the dataset, namely, SVM\cite{xu20183d},  CNN\cite{xu20183d}, and WiNN. The comparisons of three methods give the quantitative evaluation to demonstrate the efficiency of the dataset WiVi, while WiNN \cite{wang2019can} generates the skeleton results for the visual inspection of the WiVi dataset.

\begin{table}[!t]
\centering
\caption{The accuracy of three action segmentation by SVM, CNN and WiNN, respectively. PA is product accuracy. OA represents overall accuracy}
\label{tab:accu}
\resizebox{\textwidth}{!}{
\begin{tabular}{llllllll} 
\hline
\multicolumn{2}{l}{\multirow{2}{*}{\diagbox{Action}{Method}}} & \multicolumn{2}{l}{SVM}      & \multicolumn{2}{l}{CNN}      & \multicolumn{2}{l}{WiNN}      \\ 
\cline{3-8}
\multicolumn{2}{l}{}                  & PA   & OA                    & PA   & OA                    & PA   & OA                     \\ 
\hline
1s & falling                          & 1    & \multirow{9}{*}{0.95} & 1    & \multirow{9}{*}{0.98} & 1    & \multirow{9}{*}{0.96}  \\
   & throwing                         & 0.89 &                       & 0.99 &                       & 1    &                        \\
   & pushing                          & 1    &                       & 0.91 &                       & 1    &                        \\
   & kicking                          & 0.98 &                       & 0.97 &                       & 1    &                        \\
   & punching                         & 1    &                       & 0.99 &                       & 0.88 &                        \\
   & jumping                          & 0.91 &                       & 1    &                       & 1    &                        \\
   & phonetalk                        & 1    &                       & 1    &                       & 0.96 &                        \\
   & seating                          & 0.8  &                       & 1    &                       & 0.87 &                        \\
   & drinking                         & 0.96 &                       & 1    &                       & 1    &                        \\
   &                                  &      &                       &      &                       &      &                        \\
2s & falling                          & 1    & \multirow{9}{*}{0.94} & 1    & \multirow{9}{*}{0.91} & 1    & \multirow{9}{*}{0.94}  \\
   & throwing                         & 0.92 &                       & 0.97 &                       & 1    &                        \\
   & pushing                          & 0.9  &                       & 0.85 &                       & 1    &                        \\
   & kicking                          & 0.94 &                       & 0.99 &                       & 0.86 &                        \\
   & punching                         & 0.93 &                       & 0.74 &                       & 0.88 &                        \\
   & jumping                          & 0.94 &                       & 0.85 &                       & 1    &                        \\
   & phonetalk                        & 0.91 &                       & 1    &                       & 1    &                        \\
   & seating                          & 0.88 &                       & 0.89 &                       & 0.81 &                        \\
   & drinking                         & 1    &                       & 1    &                       & 1    &                        \\
   &                                  &      &                       &      &                       &      &                        \\
3s & falling                          & 1    & \multirow{9}{*}{0.97} & 1    & \multirow{9}{*}{0.95} & 1    & \multirow{9}{*}{0.94}  \\
   & throwing                         & 0.89 &                       & 0.8  &                       & 0.88 &                        \\
   & pushing                          & 1    &                       & 1    &                       & 1    &                        \\
   & kicking                          & 0.95 &                       & 1    &                       & 0.85 &                        \\
   & punching                         & 0.95 &                       & 0.87 &                       & 0.9  &                        \\
   & jumping                          & 0.95 &                       & 1    &                       & 1    &                        \\
   & phonetalk                        & 1    &                       & 1    &                       & 1    &                        \\
   & seating                          & 0.93 &                       & 0.94 &                       & 0.89 &                        \\
   & drinking                         & 1    &                       & 0.97 &                       & 1    &                        \\
\hline
\end{tabular}}
\end{table}

\subsection{Parameter setup}
\begin{figure*}[!t]
    \centering
    \includegraphics[width=\textwidth]{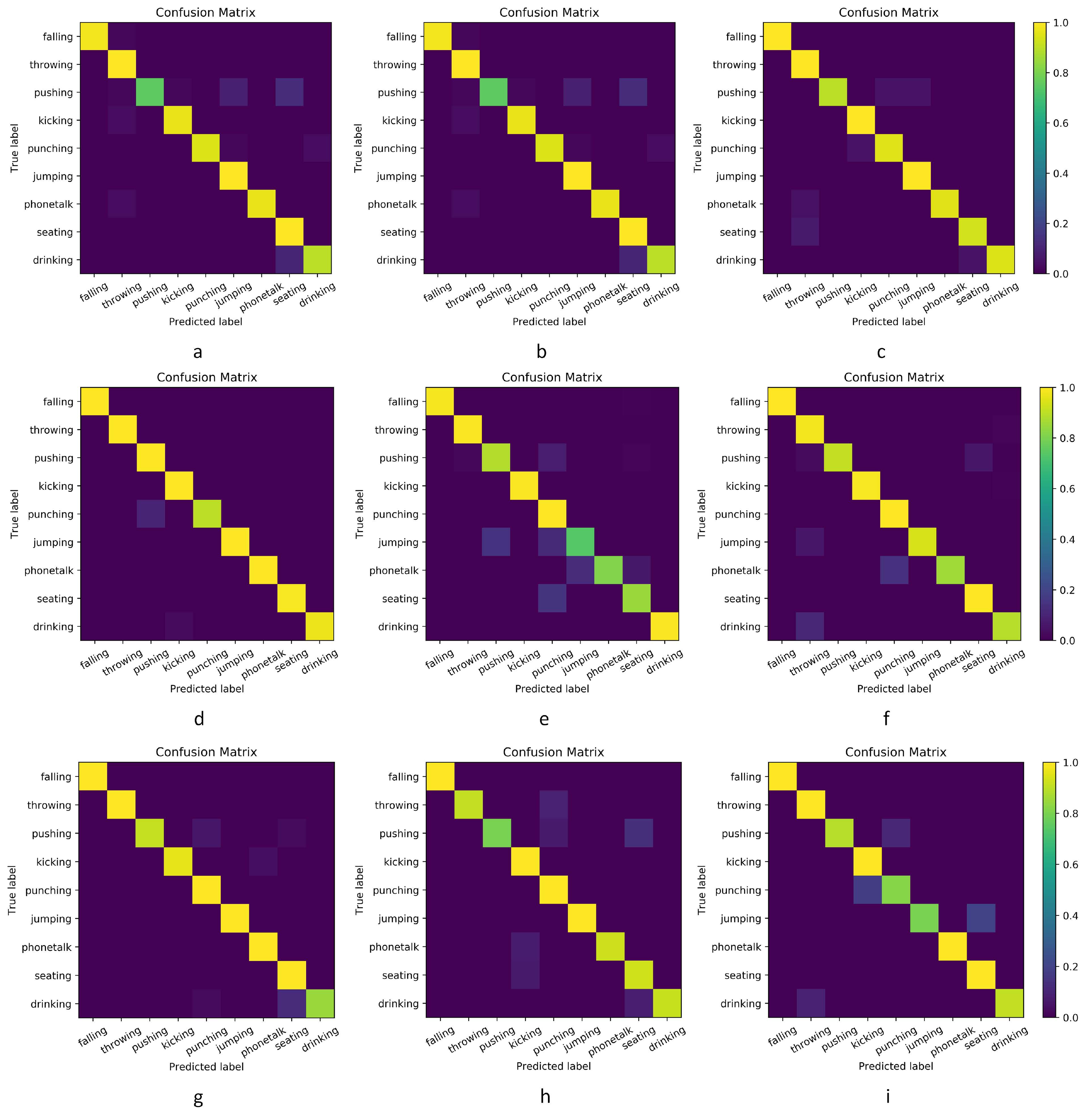}
    \caption{The confusion matrices by SVM, CNN, and WiNN in three action segmentation cases. a-c, d-f, and g-i are the corresponding confusion matrices of 1- 3 seconds action segmentation by SVM, CNN, and WiNN, respectively}
    \label{fig:cm}
\end{figure*}

\subsubsection{SVM }
SVM adopts the Radial basis function (RBF) as the kernel with a gamma of 0.143. The penalty is set as 100, and the pyramid level and probability threshold both define as  0. 
\subsubsection{CNN}
CNN is consist of 4 basic blocks of ResNet\cite{he2016deep} with 18 convolutional layers (ResNet18). At first, we upsample each CSI sample from $30 \times 3 \times 3$  into  $6 \times 224 \times 224$  to be compatible with the input of ResNet.  After this, the upsampled CSI samples are put forward to ResNet18 modules for classification.  It was trained using the Adam optimizer with a 0.001 learning rate and 0.01 decay rate for 5 epochs.
\subsubsection{WiNN}
WiNN also adopts ResNet16 where the same upsampling processing as CNN to generate the required size $6 \times 224 \times 224$. Then,  four ResNet modules are stacked together to extract the features of size  $300 \times 18\times 18$ from the upsampled CSI data. After this, two convolutional layers are constructed to decode the features into $2 \times 18\times 18$ as the skeleton features. The output skeleton features are further taken into the SVM classifier for final action classification. The parameters of SVM is the same as the SVM branch.
\subsection{Quantity Comparison}
The subsection shows the activity recognition performance using several classification algorithms on the WiVi dataset.  We analyse activity data of all volunteers. On one hand, we demonstrate the  the validity of WiVi data set; on the other, we evaluate the performance of SVM, CNN and WiNN respectively in terms of three action segmentation. 
As illustrated in Table.\ref{tab:accu}, PA is product accuracy and OA represents overall accuracy. Fig.\ref{fig:cm} shows the corresponding confusion matrices for three action segmentation by SVM, CNN, and WiNN.  

As lillustrated in Table.\ref{tab:accu}, although our data set is really challenging in terms of huge differences from scenes, persons as well as activity performances, all three methods still distinguish 9 actions well and keep more than $80\%$ of activity recognition accuracy. The results demonstrated the robustness of the data set WiVi. 

By considering each actions, our method WiNN achieves most robust results compared to  compared with SVM and CNN. For instance, WiNN obtains the most actions with full mark of classification accuracy regard to all action segmentation. To further evaluate our method intuitively , we report the results on three action segmentation in Fig.\ref{fig:cm}.

We study the impact of action segmentation on performance of activity recognition by using three classification algorithms. It shows that the most suitable action segmentation for CSI sequences varies according to the classification model. For SVM, it achieved the most accurate OA for 3s as all 9 activities classes obtain the satisfied PA to some degree. For CNN, it witnesses the highest accuracy of  $98\%$ in 1s action segmentation. Compared with SVM and CNN, WiNN achieves more robust results for 3s with respect to the PAs of 9 actions.

\subsection{Visually Interpretation}

\begin{figure*}[!t]
    \centering
    \includegraphics[width=\textwidth]{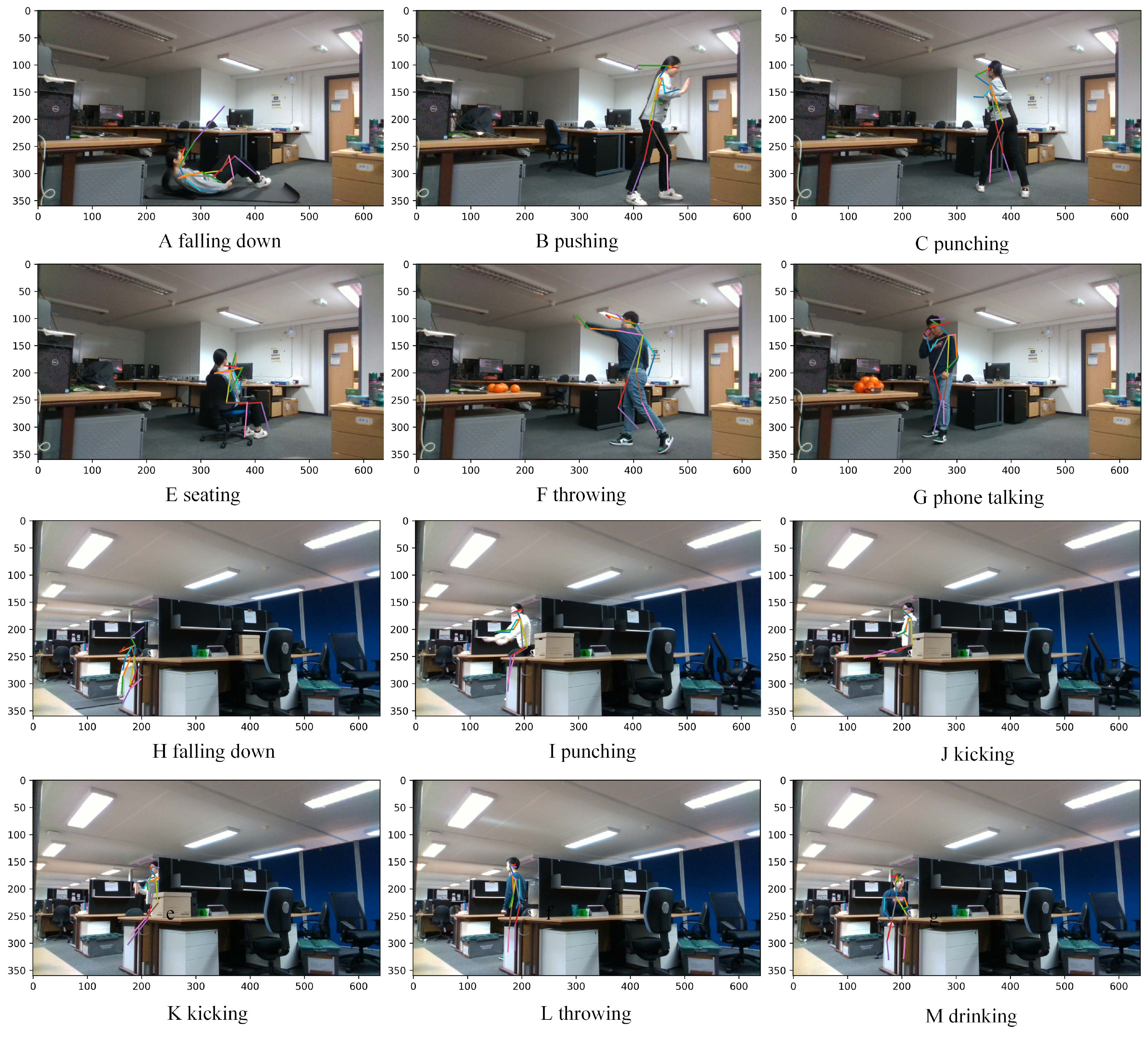}
    \caption{The visual skeleton result of the CSI  in two scenarios, where A-G are without occlusion scene, and H-M are partial occlusion scene.  }
    \label{fig:skeleton}
\end{figure*}

The section analyses and elaborates the robustness of the data set intuitively, we randomly selected twelve CSI samples and corresponding frames from each activity. To have a intuitive interpretation, we only show two scenarios, namely without occlusion scene and with partial occlusion, to produce the skeleton results by WiNN in Fig. \ref{fig:skeleton}. As shown in the figure, though the WiVi data set is diverse, it yields comparable results to videos when there is no occlusion, and even better results when there exists partial occlusion.

\section{Conclusion}
In this paper, we proposed a WiNN scheme to tackle the occlusion problem of activity recognition. To evaluate the performance of existing methods, we collected WiVi dataset as a benchmark in three scenes, including without occlusion, with partial occlusion, and with full occlusion. Three methods, namely, machine learning method SVM, deep learning methods CNN as well as the proposed WiNN were used to verify the effectiveness of the WiVi data set. Experimental results showed that WiVi dataset satisfied primary demand, and WiNN achieved the most robust results compared to other two methods regard to multiple action segmentation from 1s to 3s.  In future, it is promising to extend the diversity of WiVi dataset with respect to activity, location, and speed. Specifically, we will investigate a deep learning pipeline to realize the automatic selection of the branch for recognition according to the input data types.


\section*{Acknowledgment}

The authors gratefully acknowledge the contributions of Yixuan Zou and Zhong Yang as the volunteers of the experiments. 
 


%



  


\bibliographystyle{IEEEtran}
\bibliography{IEEEabrv,IEEEexample,mybib}

\end{document}